# CoachAI: A Conversational Agent Assisted Health Coaching Platform

Ahmed Fadhil[1], Gianluca Schiavo[2], Yunlong Wang[3]
[1]University of Trento, Trento, Italy
[2]I3, Fondazione Bruno Kessler, Trento, Italy
[3]HCI Group, University of Konstanz, Konstanz, Germany
[1]ahmed.fadhil@unitn.it
[2]gschiavo@fbk.eu
[3]yunlong.wang@uni-konstanz.de

**Abstract**
Poor lifestyle represents a health risk factor and is the leading cause of morbidity and chronic conditions. The impact of poor lifestyle can be significantly altered by individual behavior change. Although the current shift in healthcare towards a long-lasting modifiable behavior, however, with increasing caregiver workload and individuals' continuous needs of care, there is a need to ease caregiver's work while ensuring continuous interaction with users.
This paper describes the design and validation of CoachAI, a conversational agent-assisted health coaching system to support health intervention delivery to individuals and groups. CoachAI instantiates a text-based healthcare chatbot system that bridges the remote human coach and the users. This research provides three main contributions to the preventive healthcare & healthy lifestyle promotion: (1) it presents the conversational agent to aid the caregiver; (2) it aims to decrease caregiver's workload and enhance care given to users, by handling (automating) repetitive caregiver tasks; and (3) it presents a domain-independent mobile health conversational agent for health intervention delivery. We will discuss our approach and analyze the results of a one-month validation study on physical activity, healthy diet and stress management.

## Introduction

Adhering to a healthy lifestyle is among the most contributor to health promotion and disease prevention [27,28]. A varied diet and regular physical activity have significant benefits for individuals' overall health [27,28,6]. Similarly, mental wellness is associated with social competence and coping skills that lead to positive outcomes in adulthood and later stages of individuals life [29,9]. Although the benefit of pursuing a healthy lifestyle, several barriers exist in the process of health promotion. For instance, individuals' motivation to change, their demographics and preparedness are all factors that contribute to their intention to follow a healthy lifestyle. Several studies tackled the issue of poor lifestyle through mobile technologies. Approaches [31,32] developed mobile applications to mitigate the risk of poor diet, sedentary lifestyle and anxiety. That said, the learning curve associated with mobile apps is still an issue, especially for individuals with low digital literacy. According to [30], most users download zero apps per month, among which very few are used daily. Other approaches integrated wearable trackers into individuals' daily life to track caloric consumptions, steps taken and sleeping pattern [43]. However, wearable bands suffer from user abandonment in the short-, or long-term [33].

The current rise in messaging applications provided another trend to engage individuals into self-tracking their health-related interventions through a conversation medium. Text-based communication is a potentially powerful tool for behavior change, because it is widely available, inexpensive, and easy to use [44]. The boom in computational power and artificial intelligence gave the birth to messaging applications that run a conversational agent (chatbot) to handle some tasks. Such applications are already used in diet management [15,16,18], physical activity promotion [14], medication adherence [19], and mental wellbeing promotion [34]. In addition, users, including the elderly, are familiar with messaging application due to its low associated learning curve.

This paper presents CoachAI, a health intervention delivery coaching web application powered by a conversational agent and a supervised machine learning model. The machine learning model performs user clustering based on their physical exercise level. The model is based on real patient's data provided by a healthcare clinic. The platform assists the human agent with user condition tracking, provides the caregiver with insights about their users and helps track suitable user activities. The approach leverages the chatbot to levies the caregiver of menial work (e.g., data collection, reminders, patient follow-up). It provides a structured question-answer style, with a more predictable flow which caregivers can rely on. We believe the chatbot should prove to be a valuable complement to the human agent, but not a replacement. Substituting caregivers is not the aim of CoachAI, rather assisting is the purpose. In particular, we investigate how this supervised form of support is able to provide an effective engagement on users and a safe and more controlled way to train with respect to the non-tailored activities. The study will highlight the outcome of our one-month validation study with users. Our research work aims at answering the following research question: *"How different individuals' motivational factors can be integrated into the knowledge-base of CoachAI to generate support tailored to their preferences?* From our research question, we have four sub-questions:

RQ1: What is the overall user experience of using CoachAI system?
RQ2: Will user experience change over time when using CoachAI over 3 weeks?
RQ3: Will CoachAI change users' intention or behavior?
RQ4: How much do users prefer a direct human agent support or to what extent does the chatbot agent satisfy their needs?

## Related Work

Promoting individual's behavior has been shown to be quite a challenging task [41, 5]. To ease this burden while improving the care process, a study by Ibrahim et al., [1] developed a multi-agent platform to automate the process of collecting patient-provided clinical outcome measures without clinician's intervention. Health coaching approaches are widely adopted in various health domains to monitor cardiac rehabilitation [8], promote physical activity at home for elderly [13], medication adherence [19], assist pregnant women [4,7], promote healthy diet [15,16,18], support individuals with spinal cord injury [11] and assist with hand therapy [12]. Health coaching may vary in their techniques to tackle health issues. That said, most provide remote monitoring to patients through either a smartphone, sensors or wearable trackers. Health coaching is either fully automated (virtual agent), fully manual (human agent) or semi-automated (a combination of the two). Some coaching systems can be text-based [15], whereas others use an avatar or speech-powered agents [3].

Despite the abundance of health tracking and monitoring systems, tailored coaching and personalized feedback techniques are still in their infancy. A study by Villalonga et al., [2] presented an ontology-based approach to model tailored motivational messages for physical activity promotion. The ontology messages can be categorized into multiple classes e.g., sedentary, mild or vigorous activities. For example, a mild activity class is defined as *"MessageComponent $\cap$ $\exists hasAction.(Action$ $\cap$ $(\exists hasIntensity.double[>=$ "2" $double)$ $\cap$ $(\exists hasIntensity.double[<$ "5" $double]))"$. Similarly, Boratto et al., [10] provided an overview of an e-coaching system designed for runners. The platform stimulates individual's motivation to exercise provided through the coach-user interaction engagement. The results showed users tendency to be more engaged to train when their trainings are developed and remotely supervised by a human coach. The findings showed that e-coaching systems may benefit from considering the support of qualified professionals. The role of mobile apps to facilitate behavior change has shown promising results in providing rich context information including an objective assessment of physical activity level and information on the emotional and physiological state of the person [11, 14].

Current health coaching systems integrate AI-based chatbot powered with machine learning and natural language understanding [4,20,21]. Such systems are flexible and can respond to users' requests. Finally, to enhance user's engagement with health coaching system persuasive techniques and gamification design elements can be integrated within the dialog model [17, 22].

Our approach in CoachAI differs from existing literature in the technical and design aspects. Rather than relying on SMS services, we used a natural conversation to interact with users, delivered by the conversational agent. A substantial body of evidence spanning several years of research demonstrates that text-messaging interventions have positive effects on health outcomes and behaviors [44]. Most mobile apps and wearable trackers suffer from user abandonment. Instead of a standalone app or wearable tracker, we use a chatbot inside a messaging app to handle most user interactions and use the wearable tracker as a supportive tool, rather than the main tool to gather user health data. Building a free conversation-style dialog model with deep learning still has its limitations in terms of conversation accuracy and context relevancy. We relied on a task-oriented finite state machine architecture to handle the chatbot dialog. Chatbot systems are still emotionless, hence a human in the loop can cover this part and provide supports beyond the technical capabilities or in domains such as healthcare where precision is important. The presence of a human-in-the-loop avoids the shortcomings of pure agent-based control, affording an experience in which the system responds appropriately to both verbal and nonverbal cues in dialogues with a user. Self-assessment of how one handled different situation and a feedback about one's performance from a coach enhances learning from the experience.

## CoachAI: Text-based Chatbot Assisted Coaching Platform

CoachAI is a semi-automated behavior change intervention tool that relies on a conversational agent to assist the caregiver. It engages users into health-related activities to promote their lifestyle by delivering health behavior change interventions with a chatbot. The platform helps caregivers with initial baseline user's classification (profiling) and activity assignment. The platform then structurally creates a user profile by applying a supervised machine learning approach (namely, Support Vector Machine)[1] to analyze the collected data and present them to the coach. The platform is domain independent and general in terms of architecture to integrate it into any domain falling in the context of health promotion. The application uses Telegram Bot Platform[2] as the communication channel for user interaction and a web application dashboard for the caregiver to track and assign user specific activities. Using the system, the user is guided by a personalized dialogue with the chatbot. The dialogue engine is handled by a state machine, where the user switches between states until he/she fulfills the dialogue objectives. We apply data analytics to find user-dependent patterns in the link between their health parameters and the activities assigned. In addition, we aim to involve health professionals only when the user shows no adherence to the plans and no interaction with the chatbot for a period, defined by the coach. For example, when the user did not interact with or access their application for some period.

### 1. Coaching Layer

This refers to the two main layers forming the coaching portal in CoachAI, namely the coaching dashboard and the conversational agent. The coaching layer contains different components that interact with each other to form the user-chatbot interaction for the different components forming CoachAI and their interaction.

#### 1.1. Coaching Dashboard

This is the coaching panel to register and track users (see Figure-1). It contains all user's data about their daily exercise and diet, and their overall wellness coping. The dashboard allows the coach to assign initial tasks to the users and check their adherence over time. One task is formed by several plans, with each plan containing a set of activities. Moreover, the machine learning model helps the coach to cluster new users into three distinct groups. The assigned plans have an associated timeframe, performance validation, and adherence measure. The plan is assigned to the task scheduler to perform the periodic plan assignment and feedback collection. User data about steps, distances and sleeping pattern can optionally be aggregated from a tracker wore by the user. This helps access more accurate data by the coach and ease the data logging for the user. User adherence (high adherent, low adherent) are determined by following a fixed threshold and taking the average of user activity adherence over the plan duration. All the data are aggregated by the chatbot operating inside the task scheduler.

---

[1] https://en.wikipedia.org/wiki/Support_vector_machine

[2] https://core.telegram.org/bots/api

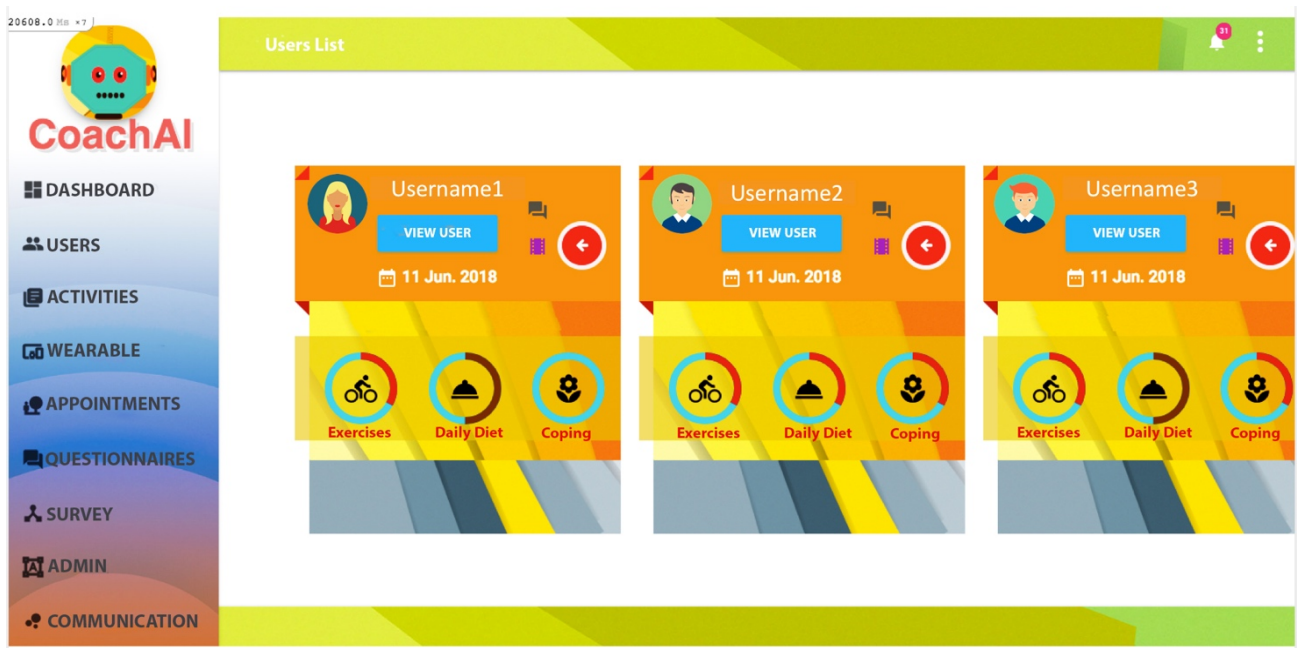

**Figure-1: CoachAI - Chatbot dashboard interface.**

### 1.2. Conversational Agent

The concept of conversational agents has already been applied to improve aspects of user lifestyle, through face-to-face conversation, using speech and hand gesture [35,36]. The conversational agent in CoachAI is based on text conversation and simple graphical elements embedded into the conversation. The user accesses and interacts with the bot on specific topics structurally defined within the finite state machine. At this step, we gather data about user's BMI, physical activity, diet, stress, and sleep. The chatbot performs two separate phases, namely information gathering and feedback collection. The chatbot tracks user's adherence in general and categorizes them into three categories, namely non-adherent, adherent, and mild users. The coach is notified in case of any user condition deteriorations (see Figure-2 for the chatbot agent). The chatbot handles tasks about activities assigned, user feedback per activity, exercises, private messages by the coach, and health intervention questionnaires.

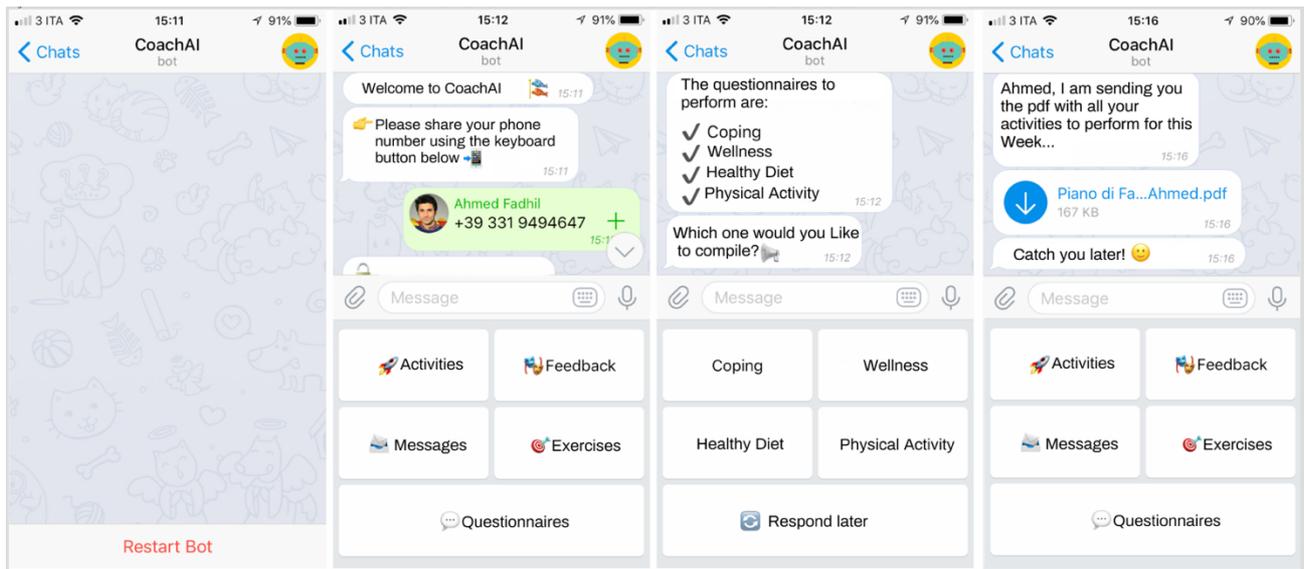

Figure-2: CoachAI - Chatbot agent interface.

2. Tasks Scheduler

The scheduler is the mechanic that steers the dialog engine with all the tasks it performs. It handles the state transition when the coach or the user are interacting with the chatbot. This may include defining a plan with the duration and duties and tracking the plan expiration date and user adherence to the plan. The task scheduler also handles all sorts of notifications between the coaching platform and the conversational agent. The scheduler notifies users when a plan has been assigned and sends them the list with the duration. The scheduler provides the appropriate plan and feedback notifications based on the time chosen for the notification. The periodic tasks are handled with a time-based background job scheduler which handles all the periodic events in the system.

3. Dialog Engine

The dialog is based on a structured finite state machine architecture that offers the capabilities to define the structure for a domain specific dialog model. Based on the defined pattern, the state machine allows constructing conversation patterns about various topics involved in the conversation. The finite state machine has unique states that guide the user through a series of steps to achieve their daily goal. We followed a standard design pattern to structure the most relevant states for health dialog models [22]. For instance, to exercise more or eat healthily, there has to be a finite set of steps a user has to fulfill to reach the end-goal of the dialog.

4. Intervention Delivery

The assessment collects data about user dietary habits, physical activity, emotional state, and more. These data help the coach build a plan and assign it to the user. The coach can also deliver standard or customized interventions through the platform. The intervention may consist of some follow-up dialogs with the user or a structured questionnaire for the user to perform to evaluate their condition. The questionnaire consists of a set of questions with their answers and a score for each answer. This approach is based on the Kaiser Pyramid model of care (KP) [37]. The user answers are scored, and a total score is given for each user responses. The platform can deliver any type of health and wellness intervention, such as cognitive behavioral therapy (CBT) [38].

5. User Classifier

We used a multi-class SVM classifier to separate new users into three distinct groups based on their initial activity level. The dataset used to train the model was anonymized user data provided by an ambulatory healthcare clinic. Based on the given data, we constructed a health questionnaire and collected data from 375 participants. The overall dataset consisted of 25 features with dependent (e.g., active or inactive individual) and independent variables (e.g., individuals' age, work, daily sedentary time, how much physical activity they perform). The independent variables cause a change in the dependent variables. The features used were mainly about user's physical activity and sedentary lifestyle. We plot each data item as a point in n-dimensional space (where n is the number of features) with the value of each feature being the value of a particular coordinate. Then, we perform classification by finding the hyper-plane that differentiates the three classes very well. Based on users' replies to the chatbot questions they are categorized into three distinct groups (namely, vigorous, mild, and sedentary users). The model considered some significant features in the data to check for when classifying the users. We used a category =3 and a response variable about the physical activity of a person. The data randomly selected samples when split into train and test.

## Methods

In this section, we define the study aim, the experiment settings, participants' recruitment and the intervention design.

### 1. Measurements

We aim to enhance user interaction with the system during all the stages of health intervention. The study investigated the design and interaction aspects, listed below, with the participants interacting. We used three models throughout the intervention to validate the dimensions listed below. The study used Health Action Process Approach (HAPA) [24] to validate the health behavior engagement. HAPA is a two-layer model: The continuum version is good for analyses and predictions, the stage version is good for interventions, hence we focus on the stage model during the experiment. The technology acceptance model (TAM) [25] was used to model how participants come to accept and use the system. Finally, the AttrakDiff model [23] was used to evaluate how participants rate the usability and design of CoachAI (see Table-1).

| Experiment Dimension | Description |
|---|---|
| **Perceived usefulness (TAM** | *This evaluates how useful the users perceived the application.* |
| **Perceived ease-of-use (TAM)** | *This evaluates how easy it was for the users to use the system.* |
| **Perceived fun (TAM)** | *This measures users' engagement and entertaining aspects while interacting with the system.* |
| **Attitude (TAM)** | *This measures users' attitude towards the system.* |
| **Intentions (TAM)** | *This measures the future aspects of users to reuse the system.* |
| **Pragmatic Qualities (AttrakDiff)** | *Pragmatic factors are, for example, usefulness and usability.* |
| **Hedonic Qualities (AttrakDiff)** | *Hedonic factors include emotional needs, such as curiosity and identification. The resulting attractiveness is based on the combination of pragmatic and hedonic factors.* |
| **Appeal (AttrakDiff)** | *This is to measure how appealing the system is to users.* |
| **Social Qualities (AttrakDiff)** | *This was added by us to check for the social qualities the system provides to the users.* |
| **The Motivation Phase (HAPA)** | *This phase is to measure user's intention to either adopt a precaution measure or change risk behaviors in favor of other behaviors.* |
| **The Volition Phase (HAPA)** | *This phase describes how hard users try and how long they persist to a behavior change.* |

**Table-1: Experiment dimensions.**

### 2. Study Design

The study initially gathered basic demographic information from participants through a text conversation with the chatbot. The information contained participants' personal data and personality test about their daily exercise, dietary pattern and stress. The intervention study lasted for one-month and participants interacted with the chatbot daily.

Participants' behavior was measured during the four-week validation study. The study contains two conditions, baseline (week 1) and intervention (weeks 2, 3, and 4). In the first week, participants were asked to chat with the bot and provide their baseline physical activity and healthy diet intention, together with their stress evaluation. The initial questionnaire and interaction with the chatbot provided an estimated data about each individuals exercise, diet and stress level. These data were considered when providing the individual plan per participant and later measured with the intervention condition (the rest of the weeks). The study focused on qualitative analysis to measure participants behavior change between the baseline and intervention condition. The overall user experience with CoachAI was measured with the validation questionnaires (TAM or AttracDiff) which was targeting RQ1, whereas their experience changes when they used CoachAI was measured over their feedback provided to the chatbot (3 weeks) which also targeted RQ2 and RQ4, and their adherence to the weekly plans was used to measure the RQ3 (HAPA).

All the data collected from participants were anonymized for the duration of the study purpose. Participants' adherence was tracked by a researcher to decide what plan or activities to suggest and to communicate with the participants when necessary. The emphasis of this study was on participants' interaction with CoachAI and trying to follow the instructions asked by the chatbot. Moreover, participants received at least a questionnaire to perform about their overall experience. There were also the private messages communicated to the participants at least once a week to report their adherence and provide them with motivational messages. The messages were fixed and tailored to each participant based on their adherence to the physical activity, healthy diet, and mental wellness activities. Assigned plans on healthy diet, physical activity and stress management were all validated and structured. All the activities assigned were based on the ChooseMyPlate [26] general recommendations and guidelines. A researcher structured the list of daily exercise, healthy diet, and mental wellness activities to be assigned to the participants.

## 3. Participants

The study involved 22 participants randomly recruited from online social networks and selected based on certain criteria. At the beginning 3 participants decided to drop out for personal reasons. The total number of participants was 19 (Male = 11, Female = 8, aged 19-53, Std. Deviation = 9.371, range = 34.00) (see Table-2), and were asked to provide their phone number and age for the registration. All subjects were healthy, with no health condition and were motivated to participate in the experiment and perform health promotion activities. Participants were either native Italian speakers or spoke Italian fluently. This was important due to the language-based nature of the study. A reimbursement of 5€ was given to all participants in the form of Amazon coupon. All participants interacted with the bot and performed the initial user profiling.

**Descriptive Statistics**

|  | Participants | Age | Gender |
|---|---|---|---|
| **Valid** | 19 | 19 | 19 |
| **Missing** | 0 | 0 | 0 |
| **Mean** |  | 28.53 |  |
| **Std. Error of Mean** |  | 2.150 |  |
| **Median** |  | 25.00 |  |
| **Std. Deviation** |  | 9.371 |  |
| **Variance** |  | 87.82 |  |
| **Range** |  | 34.00 |  |
| **Minimum** |  | 19.00 |  |
| **Maximum** |  | 53.00 |  |

Table-2: Participants' Descriptive Statistics.

## 4. Procedures

The procedure starts with the human agent registering the users, then the chatbot asks them to perform the initial questionnaire to evaluate aspects of their diet, physical activity and other health parameters (baseline). On concluding the initial questionnaire by users, the human agent receives a notification indicating user profile completion. The human coach then can either define a new health plan adherent with user condition or assigned a pre-built plan from the activity pool. All the plans have a category, a description, periodic event triggering time and expiration time. Upon plan submission, the chatbot notifies the users about the plan availability, then collects their feedback later. Overtime, the chatbot also tracks user's adherence to the plan by measuring the overall average adherence to the activities defined. For example, in a weekly plan the chatbot measures user's average adherence to all the activities. The adherence is either collected manually by the chatbot or aggregated from user wearable, in case paired with CoachAI.

At the end of each week the coach performs a follow-up with the users through the private messaging channel. The follow-up consists of a message sent to the users which contains motivational messages and a feedback on their overall compliance to the plans. In addition, 3 weekly questionnaires (namely, the TAM, the HAPA and the AttrakDiff) are sent to the users to complete. These questionnaires measure aspects of user's experience, expectations, and other interaction related parameters with CoachAI. This procedure lasted for 4 weeks and was terminated at the end of the fourth week. Figure-3 below shows the experiment setup we followed throughout CoachAI experiment design process.

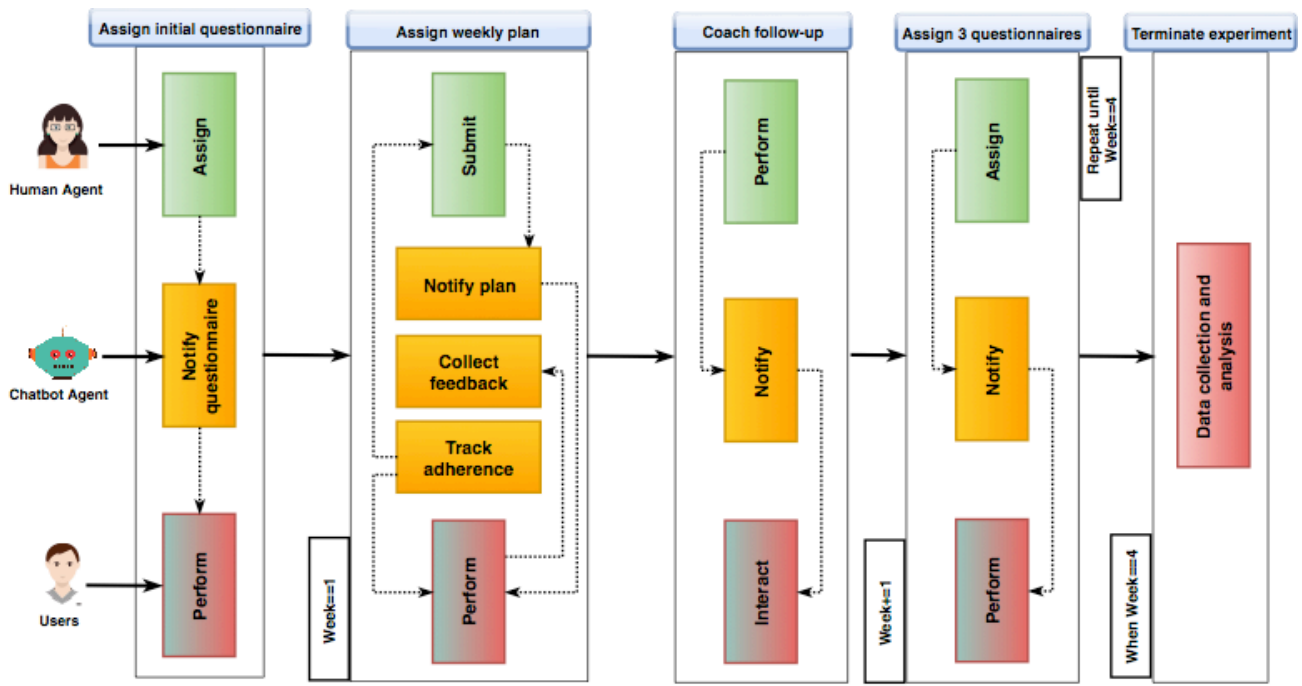

Figure-3: CoachAI experiment design process.

## Results

We analyzed the weekly data collected using the three models to evaluate users overall experience with CoachAI.

### 1. User Adherence Analysis

The system calculated participants overall adherence to the plan and reported their total adherence at the end of each plan expiration. The adherence was categorized into high and low adherence groups. The adherence measure is calculated by taking the average of all the activities performed within a plan. Users performing above the threshold were categorized as highly adherent, and vice versa.

Participants were categorized according to their adherence to the plan in two categories in high and low adherence, with respectively 10 and 8 participant each. The TAM dimensions were analyzed with a Multivariate ANOVA (MANOVA) with "Adherence level" as between subject factor. The MANOVA shows a significant effect of the between subject factor for Usefulness ($F(1,16)= 6.5$, $p<.01$), Fun ($F(1,16)= 4.5$, $p<.01$) and Attitude ($F(1,16)= 6.9$, $p<.01$). No differences for the Ease of Use, Intention. Figure-4 presents the result of the user adherence vs the TAM dimensions.

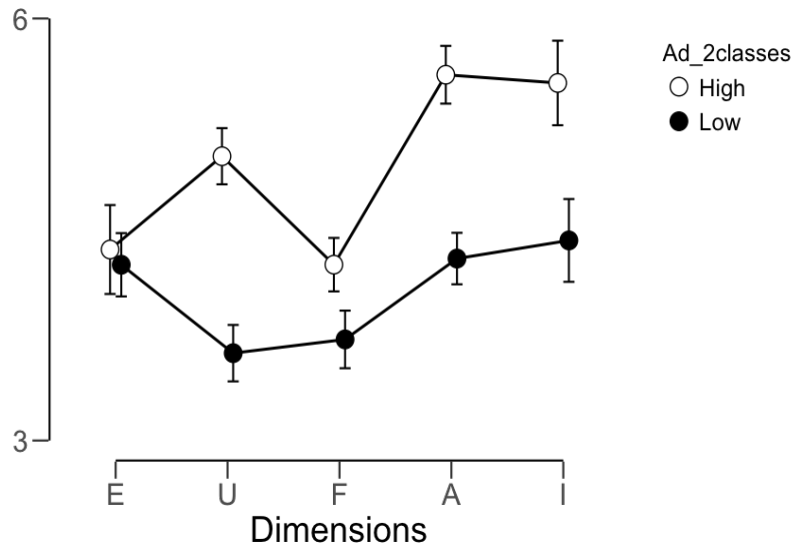

**Figure-4:** User adherence analysis and TAM. E (ease of use), U(Usefulness), F(Fun), A(Attitude), I(Intention).

## 2. User Familiarity with Chatbot Analysis

We ran a multivariate ANOVA test to measure this dimension against the AttrakDiff and TAM dimensions. Based on self-rated scale, participants were categorized by their familiarity with chatbots and the frequency of use of such technology. Participants were grouped in two categories: High familiarity (11 participants) and low familiarity (7 participants). A multivariate ANOVA was used to test for differences between these two categories with AttrakDiff and TAM dimensions. Regarding AttrakDiff, the analysis shows a significant effect of the between factor for the Pragmatic ($F(1,16)= 4.9$, $p<.01$), Hedonic ($F(1,16)= 7.8$, $p<.01$) and Appealing ($F(1,16)= 4.8$, $p<.05$) dimensions. Participants more familiar with chatbot applications scored higher in all the dimensions. Considering the TAM dimensions, the analysis shows an effect for the Usefulness ($F(1,16)= 6.2$, $p<.05$) and Fun ($F(1,16)= 8.4$, $p<.05$) dimensions. Participants with high familiarity with chatbot technology reported higher scores in these categories (see Figure-5).

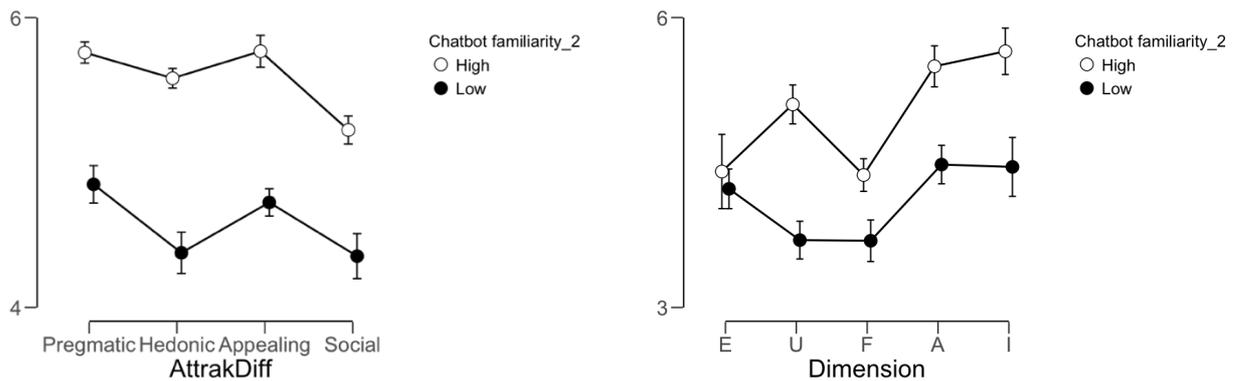

**Figure-5:** Chatbot familiarity dimension vs the TAM and AttrakDiff dimensions.

## 3. Topic specific User Interaction Preferences

To understand how the users perceived the overall CoachAI platform, we performed a self-evaluation at the end of their feedback. These self-evaluations acted as validation to user's interaction with the chatbot and measured their experience with the system when performed a topic specific activity. For example, if the user received and performed a physical activity plan, they're asked to indicate their preferences to be coached by a human agent, a virtual agent, or a combination of the two. A fixed set of validation questionnaires were asked to participants directly through the chatbot channel.

Interestingly, we found participants following a physical activity plan indicated their preferences to a virtual coach to track their physical activities. Whereas, participants in the healthy diet group preferred having a human agent to provide them with daily dietary activities and monitor their progress. Finally, participants in the mental wellness group chose a combination of a human agent and a virtual agent together to monitor, assist and support them with their anxiety and stress coping (see Figure-6 for the data). We believe this result is due to the nature of each topic which may defer in terms of needs and user expectations. In addition, some topics are more emotional bound than others and vary in the level of health complexity. For example, users following a physical activity are more interested in a tailored activity which can be easily achieved by a virtual agent, whereas those in the mental wellness group may need support beyond the technological capabilities, hence having a human in the loop will provide them with the emotional support and can change the communication tone which is crucial in such domains [15].

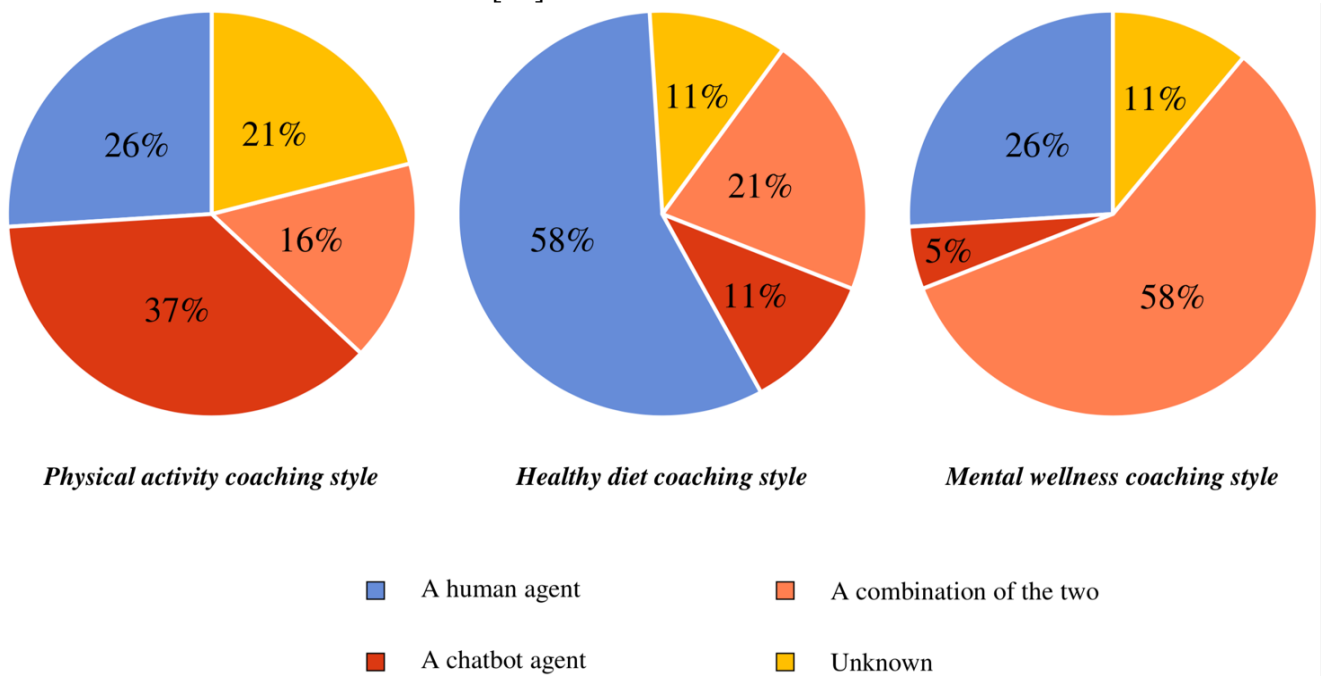

**Figure-6: Participants – Health tracking preferences by topic. Left (physical activity group), middle (healthy diet group), right (mental wellness group).**

4. Attitude and Intention

Participants highly rated both attitude and intention towards using the system. Most participants gave a rating above average to both dimensions. Considering the questionnaire on the dimensions of the Technology Acceptance Model (TAM) (see Figure-7), one-sample t-test was used to compare the means for each scale to the scale middle value (score =4). The scales "Ease of Use", "Attitude" and "Intention" are significantly higher than the middle score (respectively: $t(17)= 4.9$; $p<.01$, $t(17)= 2.5$; $p<.05$, $t(17)= 3.1$; $p<.01$). A repeated measure ANOVA (with time as a within factor) showed a difference only for the scale "Intention" between the three different weeks. Post-hoc analyses reveal that the scores for this scale at week 1 were higher than week 2 and week 3 (see Figure-8).

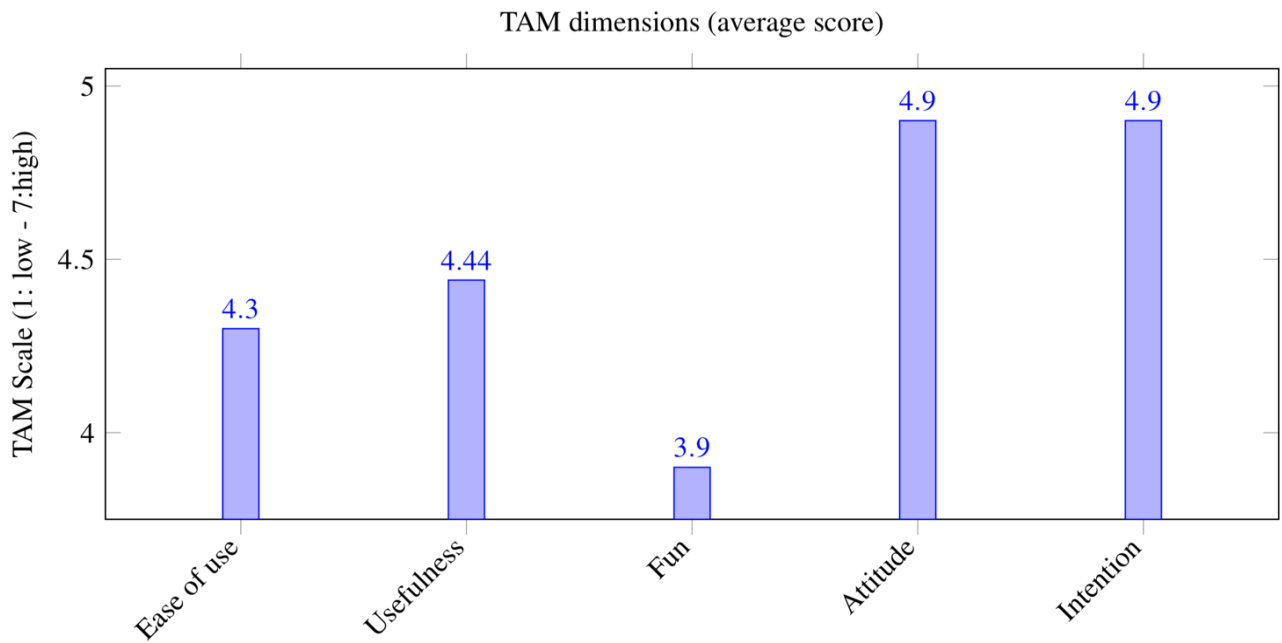

**Figure-7:** TAM dimensions (average score).

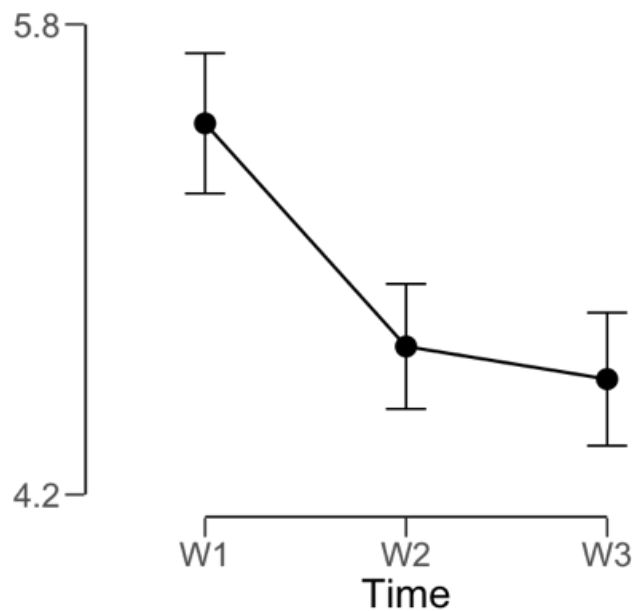

**Figure-8:** TAM Post-hoc analyses between weeks.

## 5. Pragmatic Qualities, Hedonic Qualities, Appeal and Social Qualities

This was performed over four weeks. We analyzed these qualities based on users' responses. The results of the AttrakDiff questionnaire are summarize in Figure-9 below: A one-sample t-test was used to compare the mean scores for each dimension to the scale middle value corresponding to "Neutral" (score = 4). The test shows that the average scores are statistically higher than 4 for each dimension: Pragmatic ($t(17)=5.41$, $p<.01$), Hedonic ($t(17)= 3.4$, $p <.01$), Appealing ($t(17)= 4.2$, $p <.01$), Social ($t(17)= 2.6$, $p <.05$). A repeated measure ANOVA did not find any difference between scores in the four weeks.

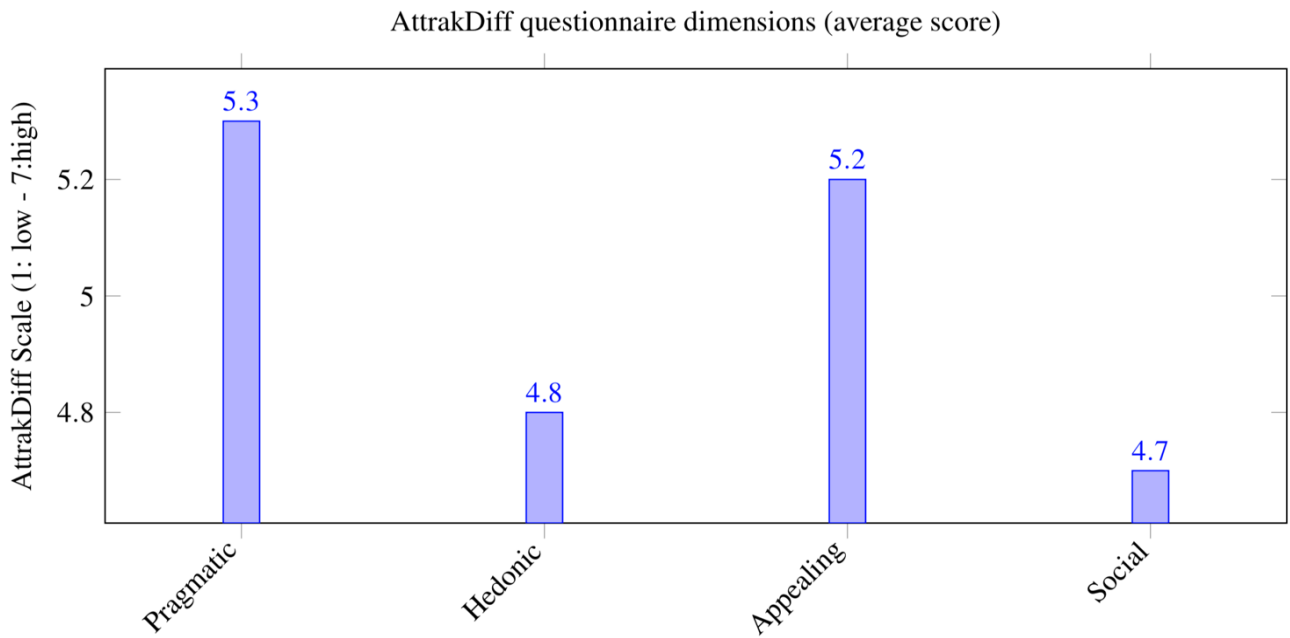

**Figure-9:** AttrakDiff questionnaire dimensions (average score).

## 6. Motivation and Volition Phase

Participants intention to change risky behavior or adopt a new measure was evaluated through the motivation and volition phases of the HAPA model [24, 42]. The HAPA Stage model was adopted to measure participants intention to change improve self-efficacy, outcome expectancy and risk perception. This model divides users into three distinct groups, namely non-intenders, intenders, and actors (see Figure-10). Users in this model can either progress their behavioral intention to change or they may also fall into relapses and recycle through the stages. We compared these stages with individual's overall adherence to the activities. A repeated measure ANOVA (with time as a within factor) showed no different between the three weeks, the scores remained unchanged. Participants were asked to provide their answers to the HAPA questionnaire over the month. We analyzed individual's data about physical activity, healthy diet and overall health expectation and mapped these data with the perceived self-efficacy, outcome expectancy and risk perception.

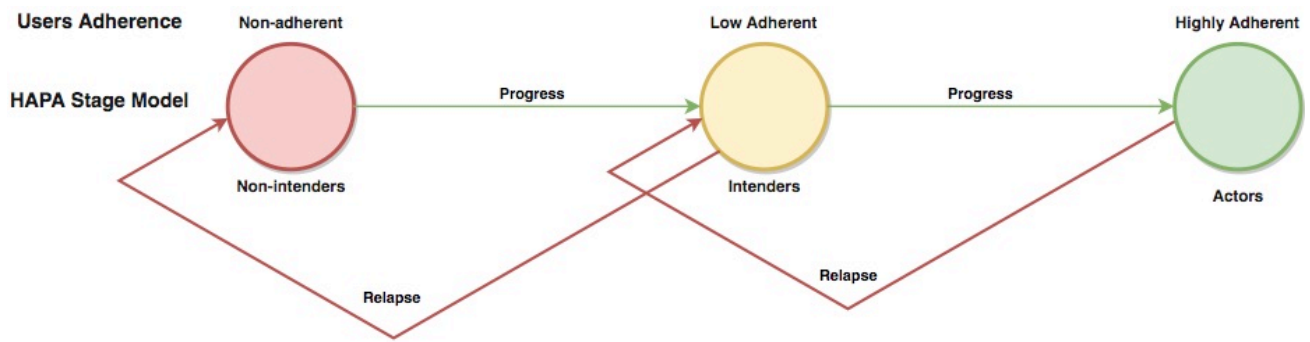

**Figure-10:** HAPA - Stage model and users' adherence.

**Physical activity intention:** Users data about their intention to pursue an active lifestyle was collected at the end of each week. The longitudinal datasets were collected through questionnaires at three points. We mapped individual's adherence and their data from the HAPA model to measure their risk perception, positive outcome expectancies, action self-efficacy, and behavioral intentions. The analysis revealed no statistically significant results when analyzed within subject and between subject effects. There was no change in individual intention throughout the three weeks.

**Healthy diet intention:** The study examined the role of self-efficacy and outcome expectancy in the context of dietary behaviors. The model includes three predictors of the intention to eat a healthy diet (action self-efficacy, outcome expectancies and health risk). A longitudinal field study was designed to examine the interrelationships of these factors with individual's adherence to healthy diet activities. The analysis revealed no statistically significant results when analyzed within subject and between subject effects.

**Mental wellness intention:** The study examined the role of self-efficacy and outcome expectancy in the context of mental wellness. The model includes three predictors of the intention to perform stress management skills (action self-efficacy, outcome expectancies and health risk). A longitudinal field study was designed to examine the interrelationships of these factors with individual's adherence to a coping skill. The analysis revealed no statistically significant results when analyzed within subject and between subject effects.

**Overall health expectation:** Users health expectancy data were collected with questionnaires at three points. We mapped individual's health expectancy data with their adherence to measure their risk perception, positive outcome expectancies, action self-efficacy, and behavioral intentions. The analysis revealed no statistically significant results when analyzed within subject and between subject effects. There was no change in individual intention throughout the three weeks. Figure-11 below shows the comparison between individuals' physical activity intention, healthy diet intention, mental wellness, health expectancy and individuals' overall adherence.

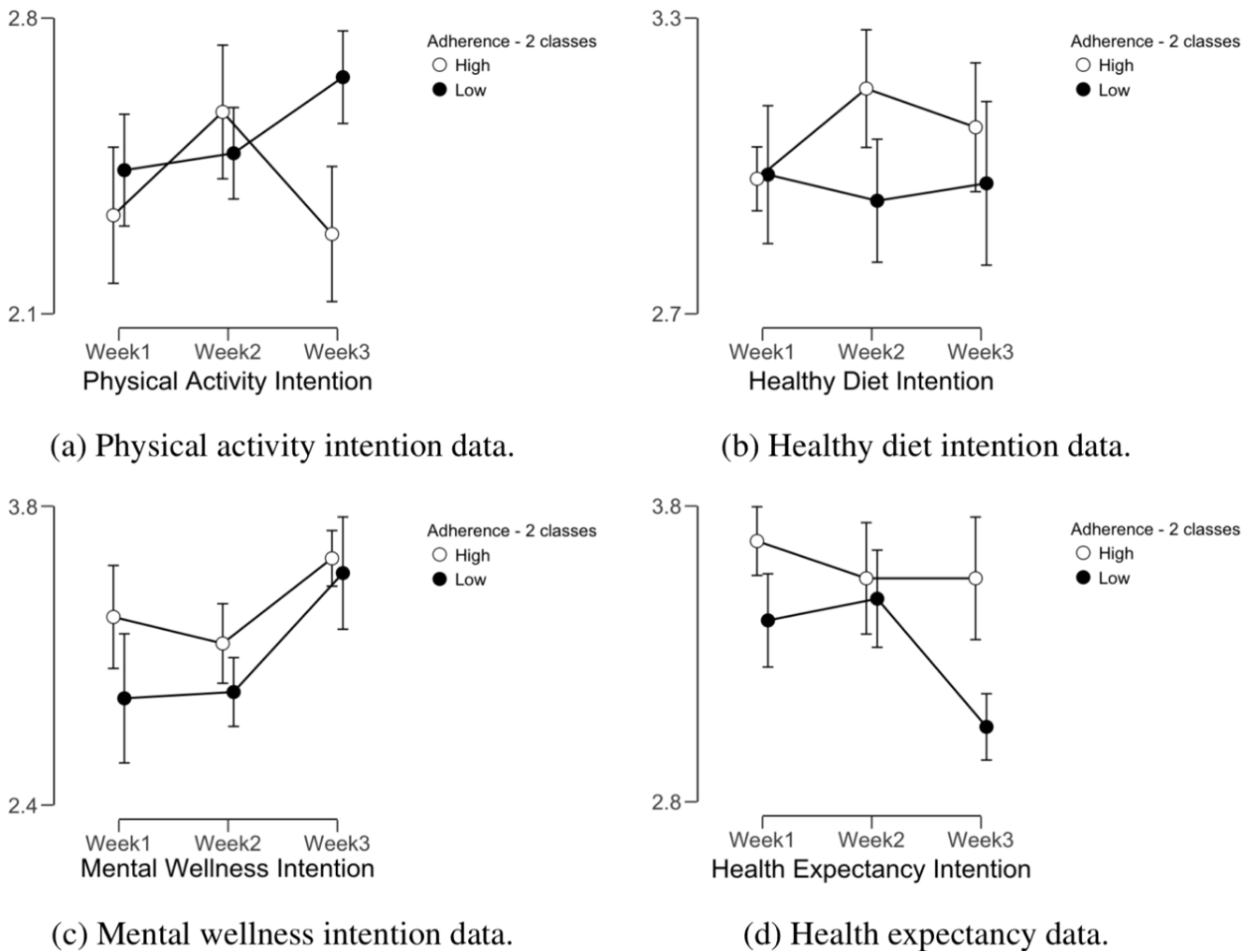

(a) Physical activity intention data.  (b) Healthy diet intention data.

(c) Mental wellness intention data.  (d) Health expectancy data.

**Figure-11: Physical activity, Healthy diet, Mental wellness, and Health expectation vs individual's adherence.**

The high and low variables refer to individual's overall adherence to the plans over the three weeks. We measured these two variables by calculating the average of user's weekly activities and dividing them by the total number of activities assigned. We compared these two variables against users' responses to each of the physical activity, healthy diet, mental wellness and health expectancy measures collected by the HAPA questionnaires.

## Discussion

Text-messaging based mobile application is a powerful tool for behavior change interventions, due to its widely availability, inexpensive, and ease of use [8,39]. A substantial body of evidence spanning several years of research demonstrates that text-messaging interventions have positive effects on health outcomes and behaviors [8,40]. CoachAI is a messaging-based conversational agent built to support the development, classification and delivery of both individual and group-based health interventions. The health activities were sent recurrently (e.g., daily, weekly, on weekends or weekdays) or at a scheduled date and time decided by the coach. Across the experiment there was strong support for the research question and the sub-questions. Overall, the experiment results clearly lead to a positive answer to the RQ1. There was some noticeable user experience emerged from using CoachAI. Users experience changed and varied while performing their activities, which confirmed our RQ2. However, there was little or no support for RQ3, user's intention towards physical activity, healthy diet, or mental wellness activities remained unchanged over the three weeks experiment. Finally, there was a strong support for RQ4, individuals performing different health activities revealed their preferences to a direct human agent support, or the chatbot agent support, or a combination of the two. The research concluded that building the technology in a user familiar environment is an effective way to increase factors of technology sustainability. The users positively perceived the activity sessions when they are planned by a human coach. Hence, the support of a human agent is clearly more motivating and engaging with respect to the virtual trainer that reveals itself alone insufficient to fulfill user's needs.

## Limitations

Several limitations worth mentioning emerged from the intervention study. First, although the activities were clinically validated and clearly guided, they were assigned by a researcher and not a healthcare provider. Moreover, participants were randomly chosen, healthy subjects and not involved in the testing as users. This was because of the study focus, which was on validating participants experience with the conversational agent. Second, a long-term follow up is required with a health professional and real patients to clinically validate the efficacy of the platform in assisting the caregiver and supporting their users. We acknowledge these as a potential limitation to the generalization of the results.

The other limitation was with the finite variability within the system which becomes less engaging because they eventually become predictable. The set of coaching activities, coaching messages and chatbot follow ups with users are renewed every time and iteration, which minimizes the repetition involved, hence decreases the finite variability in the system. Nonetheless, since the state machine steering the dialog and tasks is determined and finite, this might impact user's engagement in the long-term. There are several research questions that CoachAI can be used to address. For example, future work will investigate whether building the technology in line with users messaging habits of using messaging applications will increase the efficacy of the system, and if there is any difference between a coaching system with and without a human coach.

## Conclusion and Future Work

Existing health applications for lifestyle promotion mostly focus on handling user's condition and recommending activities and plans to improve their condition. We took a different approach and focused first on increasing the caregiver's time efficiency and decrease their fatigue during user follow up, and hence improving the efficiency of care. Conversational agent powered health coaching systems can offer a lot of benefit to the mHealth domain both for healthcare providers and patients. Chatbots cannot replace what humans are good at, but they can provide an interesting channel to support patients in delivering services through a simple conversation delivering personalized care.

Our system was evaluated in a one-month validation study, where participants interacted with the chatbot on different health related topics provided with the health coaching portal and providing them with clinically validated health activities. The study provided a set of dimensions when building chatbot powered health intervention tools. The results validated some of the questions and provided interesting insights when using conversational agents in health coaching systems. Future work will try to overcome some of the limitations emerged during the experiment. To identify patterns of use among participants, we plan to analyze the log data from the experiment. This will help understand when participants interact with the chatbot, hence identify a fingerprint for their optimal notification timing. Finally, a randomized control trial experiment with/without a human coach (with control and intervention group) will validate if users perform more activities when supervised by a human agent over the chatbot agent.

**Conflicts of Interest**

The authors declare no conflict of interest.